\documentclass[a4paper,fleqn]{cas-sc}
\usepackage[authoryear,longnamesfirst]{natbib}
\usepackage{hyperref}
\usepackage{url}
\usepackage{newtxtext}
\usepackage{ulem}
\usepackage{color}

\def\tsc#1{\csdef{#1}{\textsc{\lowercase{#1}}\xspace}}
\tsc{WGM}
\tsc{QE}
\tsc{EP}
\tsc{PMS}
\tsc{BEC}
\tsc{DE}

\begin{document}
\let\WriteBookmarks\relax
\def\floatpagepagefraction{1}
\def\textpagefraction{.001}

\shorttitle{Review on Action Recognition for Accident Detection in Smart City}

\shortauthors{Adewopo et al.}

\title [mode = title]{A Review on Action Recognition for Accident Detection in Smart City Transportation Systems}                      
\tnotemark[1,2]



%
\author[1]{Victor Adewopo}[
                        orcid=0000-0002-1700-5241
                        ]

\cormark[1]


\ead{adewopva@mail.uc.edu}

\ead[url]{https://www.linkedin.com/in/adewopo-victor/}



\address[1]{School of Information Technology, University of Cincinnati, Cincinnati, OH, United States}
\author[1]{Nelly Elsayed}[
                 orcid=0000-0003-0082-1450
                 ]
\ead{elsayeny@ucmail.uc.edu}

\author[1]{Zag ElSayed}[%
    orcid = 0000-0001-9094-1469
  ]
\ead{elsayezs@ucmail.uc.edu}



\author%
[1]
{Murat Ozer}[
  orcid = 0000-0002-2518-3398
]
\ead{ozermm@ucmail.uc.edu}

 \author%
[2]
{Ahmed Abdelgawad}[
orcid = 0000-0002-6655-2065
]
\ead{abdel1a@cmich.edu}

\address[2]{School of Engineering and Technology, Central Michigan University, MI, United States
 }
\author%
[3]
{Magdy Bayoumi}[
orcid= 0000-0002-0630-5273
]
\ead{magdy.bayoumi@louisiana.edu}

\address[3]{Department of Electrical and Computer Engineering, University of Louisiana at Lafayette, LA, United States}




\begin{abstract}
Action detection and public traffic safety are crucial aspects of a safe community and a better society. Monitoring traffic flows in a smart city using different surveillance cameras can play a significant role in recognizing accidents and alerting first responders. The utilization of action recognition (AR) in computer vision tasks has contributed towards high-precision applications in video surveillance, medical imaging, and digital signal processing. This paper presents an intensive review focusing on action recognition in accident detection and autonomous transportation systems for a smart city. In this paper, we focused on AR systems that used diverse sources of traffic video capturing, such as static surveillance cameras on traffic intersections, highway monitoring cameras, drone cameras, and dash-cams. Through this review, we identified the primary techniques, taxonomies, and algorithms used in AR for autonomous transportation and accident detection. We also examined data sets utilized in the AR tasks, identifying the main sources of datasets and features of the datasets. This paper provides potential research direction to develop and integrate accident detection systems for autonomous cars and public traffic safety systems by alerting emergency personnel and law enforcement in the event of road accidents to minimize human error in accident reporting and provide a spontaneous response to victims.
\end{abstract}


\begin{highlights}
\item Computer Vision in Accident Detection in Smart City Transportation System.
\item Action Recognition for Accident Detection in Autonomous Transportation.
\item Smart City Traffic Monitoring and Safety.
\end{highlights}


\maketitle
\section{Introduction}\label{sec1}
\label{sec:introduction}
In the field of computer vision, action recognition is a domain that has gained much attention since the advancement of convolution neural networks (CNNs) as a tool for solving complex computer vision problems and has received much attention in the research community over the past few years
~\cite{chattopadhyay2017grad}. Action recognition has been used in several real-life applications such as safety and security AI~\cite{al2020review,bo2021skeleton}, healthcare AI~\cite{muhammad2021human,gabrielli2019action}, and media AI~\cite{ren2002human,gedamu2021arbitrary}. Developing algorithms that can intuitively detect actions in video streams presents an opportunity to advance the research frontier using AI for human action recognition. Action recognition cuts across three major activities: feature detection, action representation, and action classification. Detecting actions in a sequence of images or video streams presents a unique challenge based on cluttered background, occlusion, and difficulties labeling human actions that are distinct from one person to another~\cite{Roboticsfrontier}.

Action recognition with a single action class in a video stream lags in practical applications. However, action localization in untrimmed video is a more tedious task. It involves developing architectures that can accurately set boundaries and end-to-end train algorithms that recognize action classes in an untrimmed video stream~\cite{lv2007single}. Pose estimation algorithms have been widely proposed in action recognition problem solving to help recognize and understand how each action happens~\cite{jhuang2013towards}. Pose estimation has successful results in multiple human action recognition tasks such as~\cite{yao2011does},~\cite{xiaohan2015joint},~\cite{cheron2015p}, and~\cite{raja2011joint}. However, the pose estimation has not shown significance in vehicle accident detection due to the specificity of the problem and the differences between human and vehicle actions and physical construction.

Transfer learning is a commonly used technique in extracting features of deep neural networks that have been trained on a specific domain with robust data set to a new domain/area of application with reduced computational resources. 
Previous research has leveraged the transfer learning approach to improve video stream action localization.~\cite{Iqbal2019} experimented with action localization on pre-selected frames by leveraging transfer learning from the existing model. The overarching goal was to simplify the complex architectures, expensive computation cost, and inefficient inferencing in existing methodologies. 

The current research trend in action recognition is focused on a classical deep neural network with two stream architectures (RGB and Optical flow)~\cite{sevilla2018integration}.
Transferring features from the pre-trained model on small action classes significantly improves AR models' performance, while other areas of focus have been on temporal localization and segmentation of actions in the untrimmed video. Hidden Markov's model has been used to capture long-range dependencies in frame-wise action recognition \cite{kuehne2014language}. In contrast, Spatio-temporal convolution and semi-hidden Markov model were used in capturing multiple actions transitions in untrimmed video~\cite{lea2016segmental}.~\cite{Iqbal2019} utilized the transfer learning technique with the I3D network on the temporarily untrimmed video to localize all action class instances in a video stream. Their experimental research using deep vanilla temporal convolutions network on features extracted from the I3D yielded the state-of-the-art result with a lightweight model and simple convolution network to extract features from the existing model without multiple layers and gated convolutions~\cite{Iqbal2019}.

This paper provides a comprehensive review of action recognition focusing on accident detection and autonomous transportation in smart city transportation system. This review includes the state-of-the-art techniques that researchers have proposed, taxonomies of AR tasks, AR applications domain, and transfer learning algorithms from complex architectures. In addition, we provide the potential future research questions in new application domains leveraging existing model architecture. 
The main contribution of this paper can be summarized by:
\begin{itemize}\label{list1}
    \item Providing a comprehensive comparison of different action recognition techniques and taxonomies used in smart city transportation systems and synthesizing the state-of-the-art research findings within the past ten years on autonomous transportation.
    \item Interpret and analyze the currently used datasets, algorithms, and metrics used by relevant research in the traffic control and accident detection domain.
    \item Explore literature gaps in existing methodology that can be addressed by current technological advancement.
    \item Identified potential future research questions that leverage existing methodology with reduced model complexities and computation resources.
\end{itemize}

The structure of this paper is organized in the following: Section (\ref{literature}) presents background and existing literature review on the domain mentioned above. The literature search, methodology, inclusion, and exclusion criteria are discussed in Section (\ref{Methods}). The results of our research and detailed analysis are discussed in Section (\ref{result}). Finally, Sections (\ref{limitation} and \ref{conclusion}) elaborated on the limitations and conclusion of the study.  

\section{Action Recognition Applications}\label{literature}
Action Recognition is a revolutionary topic in machine learning and computer vision that has been utilized in intelligent systems such as Human Assisted AI (e.g., surgery~\cite{sharghi2020automatic,ahmidi2017dataset}, sports~\cite{zhou2020human,davar2011domain}, education~\cite{ren2002human}), smart cities~\cite{al2019multimedia}, safety and security~\cite{dhulekar2017human,kamthe2018suspicious}, crisis informatics~\cite{yan2019research}, medical imaging~\cite{khan2014semantic}, and robotics~\cite{kruger2007meaning,rodriguez2020shedding}. Considering the wide application area of AR, in this research, we limit our scope to the application of Action Recognition addressing the accident detection in smart city autonomous transportation.

\subsection{Action Recognition in Smart City} 
A futuristic direction in computer vision tasks is the application of the intelligent system in autonomously performing human activities that are somewhat repetitive in nature and capital intensive. 

In a smart city surveillance system, violence can easily be spotted to alert appropriate enforcement agencies with automated analysis of video contents in surveillance camera~\cite{Fortun2015}.
The community-based monitoring paradigm focuses on tracking users, monitoring emergencies, and responding to them. The SenSquare system was implemented using crowd-sensing heterogeneous data sources for gathering data and developing classification algorithms in order to detect potential hazardous behavior in the environment~\cite{elsayed2021intrusion,sensquare,azumah2021deep}.
Law enforcement agencies continuously face an uphill battle in controlling the increase in crime rates, and gun violence, the deployment of intelligent surveillance cameras can assist in the automatic detection of firearms and alert security agencies in real-time when a firearm has been detected. \cite{romero2019convolutional} developed an object detection model that can detect firearms and crime scenes in dangerous situations based on Yolo's object detection framework using surveillance cameras.

Human behavior and specific human actions can be analyzed and classified using imaging and AI technologies. The application of AR models in understanding human behavior offers possibilities for smart city safety, especially in the aspect of tracking drivers' behavior. The National Highway Traffic Safety Administration (NHTSA) report highlighted an increase in the number of fatalities caused by distracted drivers between 2019 and 2020, which is higher than the number of fatalities caused by total accidents in 2017. The number of fatalities caused by distracted drivers rose to more than 8.5\% of total fatalities during 2017 \cite{stewart2022overview}. ~\cite{celaya2019texting} proposed a deep convolution neural network for detecting texting and driving behavior using a car-mounted wide-angle with a pre-trained Inception v3 model.
Emerging technologies such as the AR model can be integrated with CCTV cameras to reduce fire accidents in smart cities. As described in~\cite{avazov2021fire} on fire detection method in smart city environments using the Yolo4 algorithm, a robust model based on augmented data (different weather environments) as well as a reduced network structure demonstrated excellent performance and is highly effective for detecting fire disasters.
In this paper, we focus on accident detection from data obtained from different types of surveillance cameras used to monitor a smart city's transportation system.

\subsection{Action Recognition in Autonomous Transportation and Accident Detection}
Robotics and auto navigation has also benefited from the use of AR system for automatic guidance, specifically in obstacle detection, accident prevention, and lane departure assistance~\cite{Fortun2015}. Accident detection in autonomous transportation systems is essential for tracking vehicles and identifying anomalies in traffic patterns.~\cite{Cai2015} discussed the detection of abnormal traffic flow using clustering techniques on main flow direction vectors and a k-means clustering algorithm to identify outliers that deviates from normal trajectory pattern or motion flow in highways. Previous research has explored intelligent visual descriptions of scenes with connected image points using spatio-temporal dynamics in the Hidden Markov Model~\cite{morris2011trajectory}. More recent research work approached this challenge using machine learning algorithms and deep learning algorithms~\cite{Huang2019, Saunier2007,Robles-Serrano2021}.~\cite{Robles-Serrano2021} combined convolution layers and long short-term memory LSTM architectures in capturing spatio-temporal features from a sequence of images in video streams that have been proven to achieve better performance~\cite{lim2016speech,elsayed2019reduced} due to the capability of the convolutional layers to extract the features from each image in the video stream~\cite{kattenborn2021review} and the capability of the LSTM to learn the temporal information between images in the video sequence~\cite{greff2016lstm,elsayed2020reduced}.

Accident detection task includes the detection of the spatiotemporal dependencies in multiple frames from video surveillance. Hence correctly classifying video input as an accident is a more challenging task in developing an accident detection model.~\cite{Carreira2017} introduce a new two-stream inflated 3D ConvNet (I3D) based on a 2D ConvNet inflation. The authors seek to unravel the correlation of training on a more extensive network with performance boost by inflating the pooling kernel of pre-trained image classification architectures to an inflated two-stream inflated 3D ConvNets (I3D). The results of their proposed framework suggest that there is always a boost in performance by pre-training on a model. However, the extent of the boost varies significantly with the type of architecture.

\subsection{Accident Detection Methods}
In action recognition tasks, it has been found that many researchers propose their own datasets and evaluation criteria, making it challenging to identify the most appropriate datasets and results. Performance metrics also vary across multiple research works; developing a standardized evaluation technique will lead to more robust research in the application of Action Recognition tasks. Current methods allow some data samples to be repeated/duplicated in train/test data which directly causes bias in actual performance when evaluating a new research work~\cite{Jordao}.~\cite{stisen2015smart} examined the effects of heterogeneous devices on the final performance of the classifier on different activities using handcrafted features and employed popular classifiers such as nearest-neighbor, support vector machines, and random forest. They noticed sampling instabilities occurred across various devices.

The dataset video source also plays a significant role in designing accident detection models. Videos that are captured by a dashcam hold different data trajectories and street vision than the highway or traffic lights surveillance cameras. The dashcams capture the traffic video from a horizontal view. In such captured videos, both the camera and the surrounding objects are moving. This increases the problem complexity, especially when determining the approaching objects towards the dashcam and the objects that are approached by the car that has the dashcam itself. 
The traffic light and highway cameras record the scene in a vertical view, with the camera in a fixed position, while moving objects are recorded at a fixed point. Therefore, addressing each type of video content plays a significant role in calculating the trajectories, the acceleration of objects, and the moving directions. 


\subsubsection{Machine Learning and Statistical Models}
Most machine learning algorithms focus on vehicle trajectory, motion, acceleration, and car position in detecting car accidents.~\cite{singh2018deep} combined two algorithms using object detection and anomaly algorithm detection to identify accidents.~\cite{singh2018deep} proposed a framework that extracts deep representation using autoencoders and an unsupervised model (SVM) to detect the possibility of an accident. The vehicle's trajectories at the intersection points were used to increase the proposed architecture's precision and reliability.~\cite{joshua1990estimating} proposed mathematical relationships obtained through multiple linear and Poisson regression analyses to identify factors contributing to significant truck accidents on the highway using an accident dataset from Virginia highway traffic in combination with other geometric variables to model the percentage of trucks involved in a road accident.~\cite{arvin2019instantaneous} leveraged the availability of extensive data from interconnected devices in making correlations between erratic driving volatility and historical crash datasets from intersections in Michigan. Statistical variables such as fixed parameter, random parameter, and geographically weighted Poisson regressions, longitudinal and lateral acceleration were used in identifying road accident crash hotspots.

\subsubsection{Deep Machine Learning Models} Most deep learning algorithms focus on vehicle trajectory, motion/acceleration, and car position for detecting car accidents.~\cite{chan2016anticipating} proposed a Dynamic-Spatial-Attention (DSA) Recurrent Neural Network (RNN) for anticipating accidents in dashcam videos based on the vehicle trajectory and motion. The developed algorithm contains an objects detector to dynamically gather subtle cues and the temporal dependencies of all cues to predict accidents two seconds before they occur with a recall of 80\% and low precision of 56.14\%. The model generalizability in detecting accidents in varying weather conditions was not measured based on limited videos with rain, snow, and day/night, among other weather conditions.~\cite{Robles-Serrano2021} explored the Deep Neural Networks for accident detection using a three-stage approach by firstly segmenting visual characteristics of objects in the dataset, building on the Inception V4 model architecture to extract the temporal components of the dataset used in detecting accidents followed by the temporal video segmentation. A structural similarity index was applied to the dataset at preprocessing time to accurately select image frames within the data representing an accident or no accident as part of the temporal video segmentation to eliminate frames that do not contain event occurrence or repetition of the selected event. During the preprocessing, pixel-to-pixel comparisons were made to select a certain number of consecutive frames that contained features to train the model based on a specified threshold. Finally, the framework was designed to detect accidents automatically using Convolution LSTM (ConvLSTM) layers to capture spatial and temporal dependencies in input data~\cite{shi2015convolutional,elsayed2018empirical}. This type of neural network has been proven to have a better performance compared to the LSTM and CNN architectures when dealing with datasets that have both spatial and temporal structures. One of the potential limitations is a model bias based on vehicle types and other environmental conditions such as vehicle variety and the absence of pedestrians and cyclists.

\subsubsection{Social Network and Geosocial Media Data} The enormous amount of information being constantly shared daily across various social media platforms contains artifacts that can be analyzed to generate meaningful insights for traffic events~\cite{rashidi2017exploring}. Although the ability to monitor and analyze the exploding information manually seems impossible based on the high volume and unstructured formats of the information being presented~\cite{adewopo2022deep}. Monitoring traffic-related information on social media has been proven to be beneficial in detecting traffic events.
~\cite{xu2018sensing} provided a synthesis of research work that explored the usage of geosocial media data for detecting traffic events. Events such as road accidents, road closures, and traffic conditions are typically shared among a network of people through social media platforms. Such events can be tracked down with the aid of GPS in getting first responders to the event location and also often contain information that triggered such events.~\cite{xu2019traffic} utilized Twitter data for mining and filtering noisy data by association rules among words related to traffic events. The proposed framework achieved 81\% accuracy in classifying data into non-traffic events, traffic accidents, roadwork, and severe weather conditions. Similarly,~\cite{Salas2017TrafficED} developed a framework leveraging social media data to crawl, process, and filter social media data for implying traffic incidents and real-time detection of traffic events with text classification algorithm~\cite{gu2016twitter}.


\section{Literature Search}\label{Methods}
The literature search process, which has been performed, consists of four steps, including i) selecting eligibility criteria (Inclusion and Exclusion criteria), ii) formulating research objectives, iii) identifying search strategy, and iv) data extraction~\cite{harris2014write,wright2007write}. 
This study employed the systematic review methodology to address the research questions posited through a systematic and replicable process~\cite{gough2017introduction}. Specifically, the Preferred Reporting Items for Systematic Reviews and Meta-Analyses (PRISMA) Statement was used as a model for this review~\cite{page2021prisma}. Based on the established eligibility criteria, the papers selected were analyzed and synthesized to address the research questions postulated in the following subsection.

\subsection{Research Questions and Objectives}\label{Rqs}
Developing an AR model for specific tasks will enhance the use of AI systems in automating human actions and autonomously detecting actions in live feeds. Once the inclusion selection process has been carried out, based on pre-established criteria, the main results of the selected works are codified and extracted in order to synthesize and guide this research, the following research questions will be addressed in this study:
\begin{itemize}
    \item \textbf{\textit{RQ1:}} What are the main Action Recognition techniques/application in accident detection and autonomous transportation?
    \item \textbf{\textit{RQ2:}} What are the main taxonomies and algorithms used in Action Recognition for accident detection and autonomous transportation?
    \item \textbf{\textit{RQ3}:} What are the main datasets, features, and metrics used in Action Recognition  for accident detection task?
\end{itemize}

\subsection{Selecting Eligibility Criteria}
This review includes the research articles related to Action Recognition. It includes topics in (Autonomous Transportation, Traffic Control, and Accident Detection using computer vision published in peer-reviewed journals and published between 2012 and 2022. Based on the continuous evolving advancement in the technical field, the articles selected were published ten years before this review. Only research articles published in the English language were used.
The inclusion and exclusion criteria were detailed in Subsection~\ref{inclusion} and Subsection~\ref{exclusion}, respectively. This systematic review is based primarily on computer vision tasks using the AR model in autonomous transportation and smart city accident detection for further clarifications.

\subsection{Inclusion Criteria} \label{inclusion}
The publications needed to meet the following characteristics in order to be included: 
\begin{enumerate}
\item Articles should be in Action Recognition and Computer Vision domain.
\item Studies with validation of the proposed techniques.
\item Published within 2012-2022 (10 Years).
\item Full papers that are peer-reviewed.
\item Contains Video/Motion analysis.
\end{enumerate}
\begin{figure}[h!]
\centering
\includegraphics[scale=0.5]{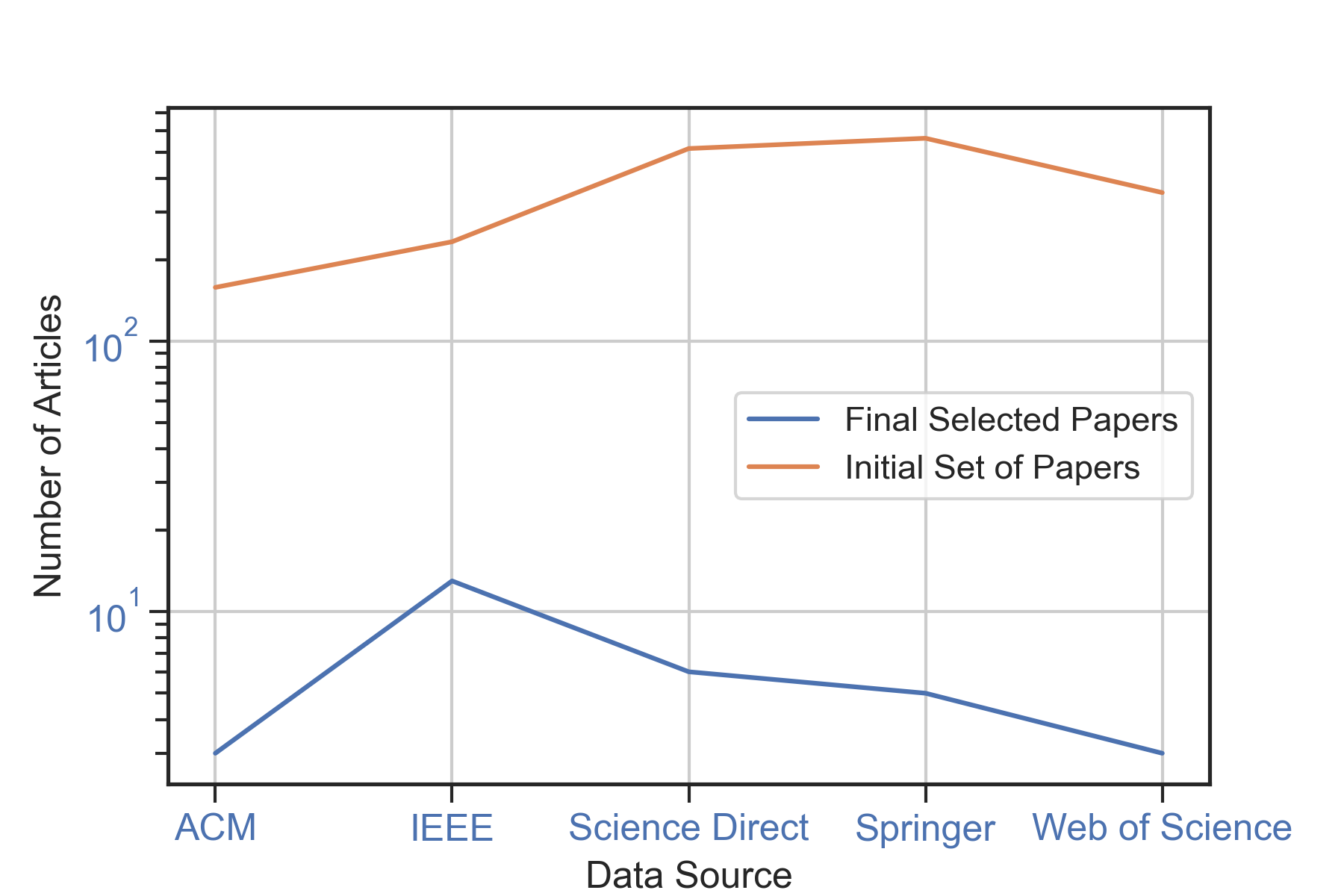}
\caption{Proportion of selected studies}
\label{selected}
\end{figure}
\subsection{Exclusion Criteria}\label{exclusion}
The following exclusions were implemented:
\begin{enumerate}
\item Does not contain video/motion analysis.
\item Published before 2012.
\item Not peer reviewed or does not provide clear findings and analysis of results.
\item Written in other languages excluding English.
\item Duplicated studies.
\item Non peer-reviewed paper.
\end{enumerate}

\subsection{Information Sources}
The papers included in this review were identified by searching electronic databases published in English.
The databases in Table~\ref{database} were used as the primary source of articles for this review.
\begin{table}
	\caption{Article Data source}
	\label{database}
	\footnotesize
	\begin{tabular}{l|l|l|ll}
		\hline
		\textbf{ID} & \textbf{Database}                      & \textbf{Link} & \textbf{Number of Articles}& \\ \hline
		D1          & \cellcolor[HTML]{C0C0C0}IEEE Xplore    & { http://ieeexplore.ieee.org/} & 299 &  \\ \hline
		D2          & \cellcolor[HTML]{C0C0C0}ACM            & { http://dl.acm.org/} & 181 &  \\ \hline
		D3          & \cellcolor[HTML]{C0C0C0}Web of Science & {
			http://www.webofscience.com/} & 445 &  \\ \hline
		D4          & \cellcolor[HTML]{C0C0C0}Springer Link & { http://link.springer.com/} &  572&  \\ \hline
		
		D5          & \cellcolor[HTML]{C0C0C0}Science Direct & {http://sciencedirect.com/} & 533 &  \\ \hline
	\end{tabular}
\end{table}

These databases provide impactful articles from full-text journals and conferences relevant to Action Recognition tasks in
smart city automation, autonomous transportation, and accident
detection.  
The first phase includes searching the databases in Table~\ref{database} with advanced search and filtering techniques to limit search results to only relevant studies. Two teaching assistants did a manual review of the search results in the second phase to ensure their validity of the search results. The number of articles retrieved from each database and the final number of papers selected is showcased in Figure~\ref{selected}. Only accessible articles are included in the search result. More details of the search term and strategy for validating and selecting relevant materials are discussed in Subsection~\ref{strategy}.

\subsection{Search Strategy} \label{strategy}
Combining the following keywords with conjunctions ``AND" and disjunctions ``OR" resulted in a total of 2,030 papers in an automated search, as shown in Table~\ref{database}. The most common terms used for our search were:
\begin{enumerate}
	\item Action Recognition.
	\item Transportation.
	\item Traffic control.
	\item Accident Detection.
\end{enumerate}

\noindent
The results of our search and the corresponding query that has been used are as follows:
\begin{itemize}
	\item {\textbf{\textit{IEEE Xplore:}}} We received~\textit{299 papers} from IEEE using the search string:[(("All Metadata":Action Recognition ) AND ("All Metadata":Transportation) OR ("All Metadata":Action Recognition ) AND ("All Metadata":Traffic) OR ("All Metadata":Action Recognition ) AND ("All Metadata":Accident Detection))] between 2013-2022
	
	\item {\textbf{\textit{ACM:}}} We received~\textit{181 papers} from ACM using the search string: [AllField:("Action Recognition") AND AllField:("Transportation") OR AllField:("Action Recognition") AND AllField:("Traffic") OR AllField:("Action Recognition") AND AllField:("Accident Detection")]
	
	
	\item {\textbf{\textit{Web of Science:}}} We received 445 papers from Web of Science using the search string:[((ALL=(Action Recognition) AND ALL=(Transportation OR Traffic OR Accident Detection))) AND (PY==("2022" OR "2021" OR "2020" OR "2019" OR "2018" OR "2017" OR "2016" OR "2015" OR "2014" OR "2013" ))]
	
	\item {\textbf{\textit{Springer Link:}}} We received \textit{572 papers} from Springer Link using the search string: [("Action Recognition") AND (("Transportation") OR ("Traffic") OR ("Accident Detection"))] between 2013-2022
	
	\item {\textbf{\textit{Science Direct:}}} We received \textit{533 papers} from Science Direct using the search string: [("Action Recognition" AND 'Transportation')  OR ("Action Recognition" AND 'Traffic')  OR ("Action Recognition" AND 'Accident Detection')] between 2013-2022
	
\end{itemize}

\subsection{Study Selection}

\begin{figure}[h!]
	\centering
	\includegraphics[scale=0.325]{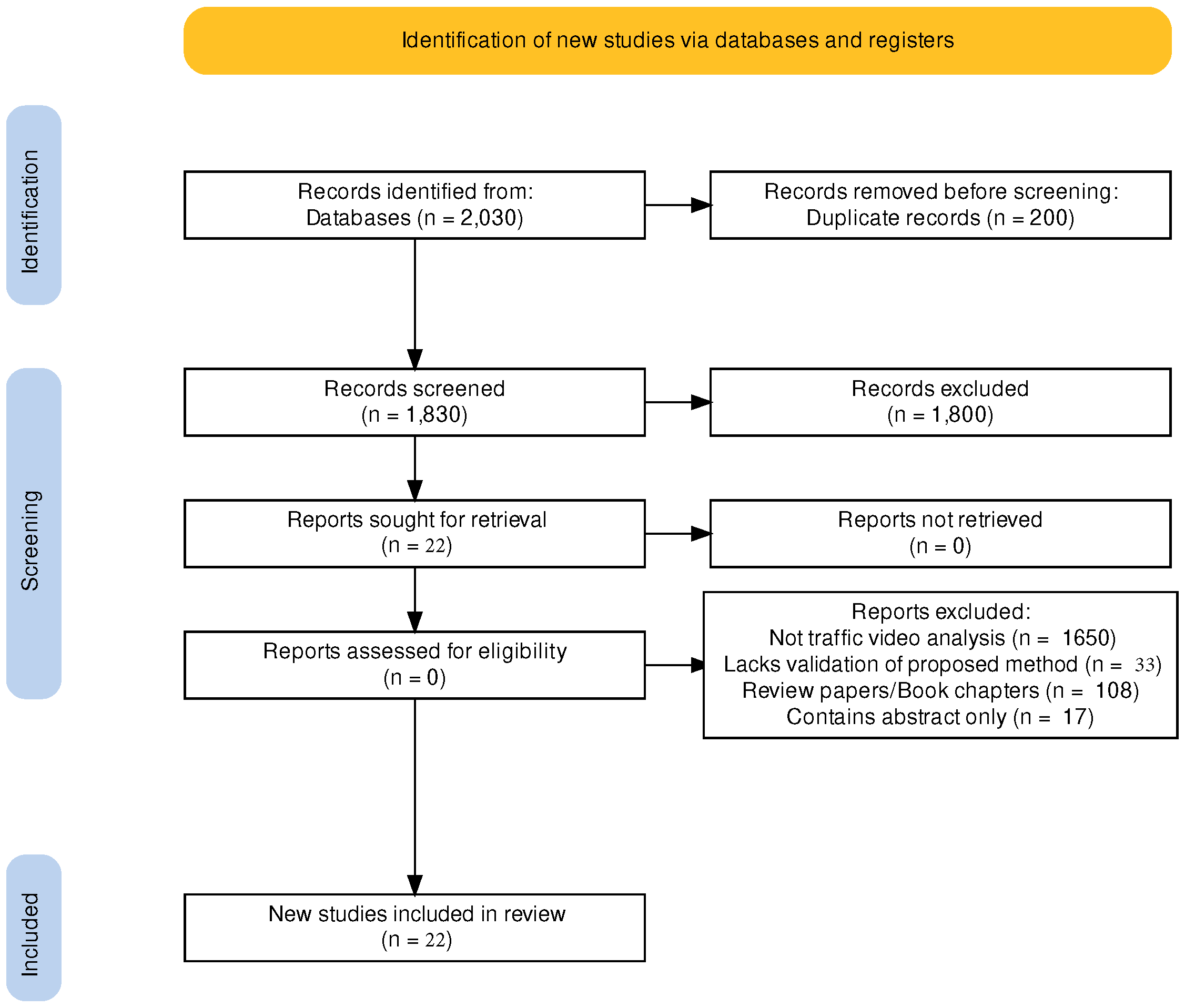}
	\caption{Preferred Reporting Items for Systematic reviews and Meta-Analyses (PRISMA) flow chart of the systematic review.}
	
	\label{Reference: Methodology}
\end{figure}

The articles were evaluated and selected according to the mentioned criteria in Section (\ref{Methods}). After the preliminary database search using the approved search strategy conducted by the student researchers and eliminating duplicates, a total of 1830  articles were screened by two faculty researchers and one student researcher independently who are domain experts. The abstracts, titles, and keywords from selected articles were reviewed for relevance based on the inclusion and exclusion criteria. Articles that did not meet the eligibility criteria or were not relevant to address the research question were removed. The independent researchers rated each article based on the inclusion criteria and eligibility criteria. The painstaking protocol observed in the selection process ensures that all articles included are relevant to this study.
A total of 1650 papers were excluded because they do not contain video analysis or employ AR techniques in detecting accidents. Thirty-three papers were excluded because they lacked validation techniques for the proposed methodology, 108 papers identified as review papers were excluded, and 17 papers contained only abstracts. Finally, only 22 papers were selected for analysis, as shown in Figure~\ref{Reference: Methodology}.

\subsection{Coding, Data Extraction and Analysis}
For the data extraction phase, the full text of chosen paper was shared among the authors for review and tagging the key contributions. Microsoft spreadsheet~\cite{niglas2007media}, Airtable~\cite{dirk2018archives} and Mendeley~\cite{zaugg3610creating} citation manager were used to coordinate workflow and analyze the papers.
This research aims to retrieve action recognition research articles relevant to accident detection and autonomous transportation. In addition, duplicate studies that cover the same issues are excluded from the study. Figure~\ref{selected} shows the proportion of initial articles and final articles selected from each of the five online data sources listed in Table~\ref{database}.

\section{Results}\label{result}
\begin{figure}[h!]
	\centering
	\includegraphics[scale=0.5]{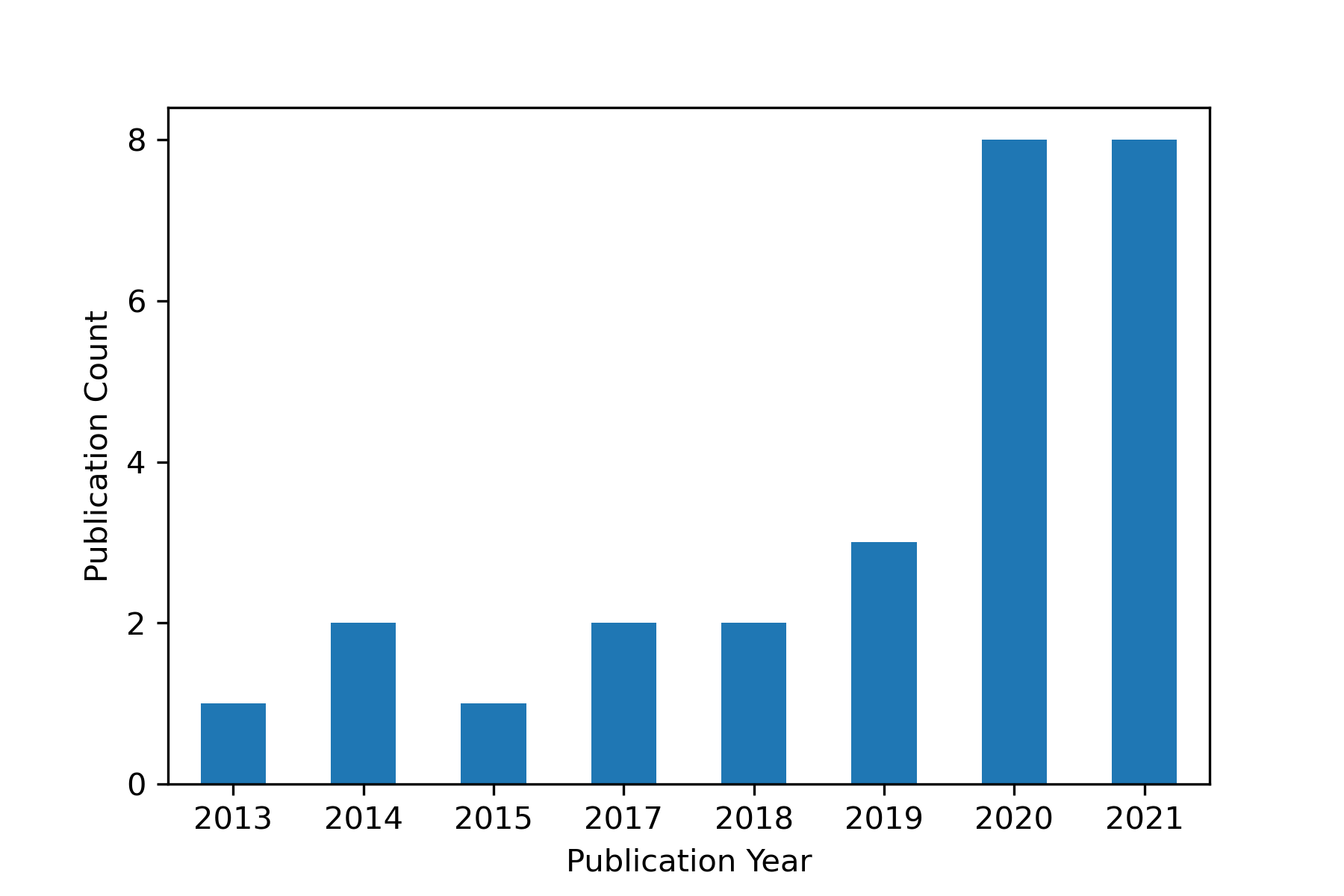}
	\caption{Number of papers published per year surveyed.}
	\label{pubyear}
\end{figure}

Following PRISMA guidelines, 2030 publications were identified through the five databases included, and the results of the 22 papers selected for review are presented in this section. 
Figure~\ref{pubyear} showcases the publication year for the selected papers. It is noteworthy that the majority of the selected papers were published between 2019 and 2021. Taking advantage of the advancement in technology and smart city automation, more research is now being conducted on which deep learning algorithms are developed to model traffic-related activities in a smart city utilizing computers equipped with high-performance GPU processors.

\subsection{RQ1: Main Action Recognition Techniques in Smart City Transportation}

The first research question of our study is to examine the main AR techniques and applications within smart cities and autonomous transportation, as shown in Table~\ref{Rqs}. Many researchers have proposed other methods to model traffic management and traffic prediction, including the use of Vector Auto-Regression, Support Vector Regression, Auto-Regressive Integrated Moving Average (ARIMA), Kalman Filter, and, most recently, LSTM, and RNN \cite{smola2004tutorial,Wang2020a}. In time series data, such as traffic control data, the approaches have not been able to capture both spatial and temporal information concisely. Recent efforts have improved the accuracy performance of GNN and GaAN~\cite{zhang2018gaan,zhao2019t}.~\cite{Ijjina2019} proposed a supervised deep learning framework to detect and identify road-side vehicular accidents by extracting feature points based on local features such as trajectory intersection and velocity by detecting anomalies in real-time accident conditions such as daylight variations.~\cite{Fernandez-Llorca2020} study utilized a visual cue derived from a camera to detect lane change or vehicle maneuvers by utilizing a disjoint two-stream convolutional network and a spatiotemporal multiplier network.~\cite{You2020} discovered that time segmentation methods such as SS-TCN and MS-TCN were more successful at higher IoU thresholds. Their experiment also suggests that the R-C3D algorithm has a comparable result when compared to segmentation-based approaches. Although newer methods like R(2+1)D and SlowFast have improved accuracy, most techniques fail to capture traffic anomalies accurately on DoTA datasets, suggesting the problem of traffic anomaly classification is challenging.~\cite{Yao2022} suggest that distant anomalies and occluded objects are difficult to classify because of their low visibility. Collisions with moving vehicles present a similar problem since, at times, the vehicle ahead is substantially obscured by the vehicle it impacts. 
There may be instances when a vehicle hits obstacles that are not detected, such as bumpers or traffic cones. Most often, anomalous vehicles are responsible for occluding the obstacles. It is hard to detect horizontal vehicle collisions due to their slow vertical trajectory, making the anomaly subtle and thus hard to detect. The JSM-based method extracts motion trajectory to evaluate traffic scenes but ignores events that occurred in an unusual manner \cite{Xia2018,Yao2022}.~\cite{Srinivasan2020} developed a scalable algorithm for high-speed object detection (DETR), with a less complex architecture and a higher level of accuracy
compared to other object detection algorithms that are based on correlations between all objects in the video data. Table~\ref{tab:my-table} addresses the research question on main Action Recognition techniques and application in autonomous transportation. The notation ``-" indicates that the corresponding research paper did not address our research question.

\begin{table*}[]
\caption{Studies were used to address the research question on main Action Recognition techniques and application in autonomous transportation. The notation ``-`` means the research paper did not address our research question.}
\label{tab:my-table}

 \renewcommand*{\arraystretch}{2}
\resizebox{\textwidth}{!}
{%
\begin{tabular}{llllllll}
\hline
\rowcolor[HTML]{C0C0C0} 
\multicolumn{1}{c}{\cellcolor[HTML]{C0C0C0}\textbf{Article}} &
  \multicolumn{1}{c}{\cellcolor[HTML]{C0C0C0}\textbf{Actions Considered}} &
  \multicolumn{1}{c}{\cellcolor[HTML]{C0C0C0}\textbf{Learning Method}} &
  \multicolumn{1}{c}{\cellcolor[HTML]{C0C0C0}\textbf{Category}} &
  \multicolumn{1}{c}{\cellcolor[HTML]{C0C0C0}\textbf{RQ1}} &
  \multicolumn{1}{c}{\cellcolor[HTML]{C0C0C0}\textbf{RQ2}} &
  \multicolumn{1}{c}{\cellcolor[HTML]{C0C0C0}\textbf{RQ3}} &
  \multicolumn{1}{c}{\cellcolor[HTML]{C0C0C0}\textbf{Key Notes}} \\ \hline
\cite{Yao2022} &
  \begin{tabular}[l]{@{}l@{}}Vehicle Collision\\Car Trajectory\end{tabular} &
  Unsupervised Learning &
  Traffic Anomaly Detection &
  \checkmark &
  \checkmark &
  \checkmark &
  The researchers developed a benchmark dataset to assess the quality of traffic accident   detection and anomaly detection for nine action classes \\ \hline
 \cite{Yu} &
  \begin{tabular}[c]{@{}l@{}}Vehicle Collision\\ Road Condition\\ Traffic Speed\end{tabular} &
  Deep Learning &
  Accident Prediction &
  \checkmark &
  \checkmark &
  \checkmark &
  \begin{tabular}[c]{@{}l@{}}Traffic accidents can be caused by many factors, including driver behavior,   weather conditions, traffic flow, and road structures.\\      The authors investigated spatial-temporal relationships on heterogeneous data to develop a road-level accident prediction system.\end{tabular} \\ \hline
 \cite{Wang2020a} &
  \begin{tabular}[c]{@{}l@{}}Traffic Flow \\ Traffic Speed\end{tabular} &
  Graph Neural Network &
  Traffic Flow Pattern &
 \textbf{-}&
  \checkmark &
  \checkmark &
  \begin{tabular}[c]{@{}l@{}}The goal of this project is to develop a framework for analyzing stationary time series traffic data. In addition, it is able to predict traffic information\\ with a 14.1\% improvement in MAPE compared to other baselines.\end{tabular} \\ \hline
\cite{Bao2020} &
  \begin{tabular}[c]{@{}l@{}}Car Trajectory \\ Inconsistent Motion\end{tabular} &
  \begin{tabular}[c]{@{}l@{}}Graph Convolution Network\\ Bayesian Neural Network\end{tabular} &
  Traffic Anomaly Detection &
  \checkmark &
  \checkmark &
  \checkmark &
  \begin{tabular}[c]{@{}l@{}}Using GCN and BNN, the developed model can handle the challenges of relational feature learning   and uncertainty anticipation from video data \\ to anticipate an accident in 3.53 seconds with an average precision of 72.22\%.\end{tabular} \\ \hline
\cite{Reddy2021} &
  \begin{tabular}[c]{@{}l@{}}Traffic Signals,\\ Car distance limit\\ Traffic Speed\end{tabular} &
  Deep Learning &
  Accident Detection &
  \checkmark &
  \checkmark &
 \textbf{-}&
  \begin{tabular}[c]{@{}l@{}}The research investigates how to extract road features relevant to the trajectory of an   autonomous vehicle from real-world road conditions using \\ Deep Q-Learning in a real-world environment setting.\end{tabular} \\ \hline
\cite{Fernandez-Llorca2020} &
  \begin{tabular}[c]{@{}l@{}}Driver Intention Detection\\ Lane Maneuver\end{tabular} &
  Deep Learning &
  Traffic Anomaly Detection &
  \checkmark &
  \checkmark &
  \checkmark &
  \begin{tabular}[c]{@{}l@{}}This study utilized a visual cue derived from a camera to detect lane change / vehicle maneuvers by utilizing a disjoint two-stream convolutional\\  network and a spatiotemporal multiplier network.\end{tabular} \\ \hline
\cite{Ali2022} &
  Traffic Flow &
  Graph Convolution Network &
  Traffic Flow Pattern &
  \checkmark &
  \checkmark &
 \textbf{-}&
  \begin{tabular}[c]{@{}l@{}}The authors propose a hybrid model combining GCN and DHSTNet that is effective in forecasting short-term traffic patterns in urban areas in\\  order to improve traffic management.\end{tabular} \\ \hline
\cite{Wang2020} &
  \begin{tabular}[c]{@{}l@{}}Car Detection \\ Accident Detection\end{tabular} & Graph Convolution Network &
  Accident Detection &
  \checkmark &
  \checkmark &
 \textbf{-}&
  The researchers developed a new dataset and proposed a method for safety prediction. \\ \hline
\cite{Alkandari2015} &
  \begin{tabular}[c]{@{}l@{}}Traffic Flow\\ Traffic Pattern\end{tabular} & Fuzzy Logic Technique
   &
  Accident Detection &
  \textbf{-} &
  \checkmark &
  \textbf{-} &
  \begin{tabular}[c]{@{}l@{}}The main aim of this   study is to develop a methodology for controlling the length of time that a   vehicle stays in traffic based on the flow \\ of traffic and congestion.\end{tabular} \\ \hline
\cite{Riaz2022} &
  \begin{tabular}[c]{@{}l@{}}Traffic Speed\\ Car Distance\\ Accident Detection\end{tabular} &
  Deep Learning &
  Traffic Anomaly Detection &
  \checkmark &
  \checkmark &
  \textbf{-} &
  This study implemented the FWPredNet framework for accident and anomaly prediction,  which outperformed the previous state-of-the-art framework. \\ \hline
\cite{Wang2020a} &
  \begin{tabular}[c]{@{}l@{}}Car Trajectory\\ Car Position\end{tabular} &
  Probabilistic &
  Traffic Flow Pattern &
  \checkmark &
  \textbf{-} &
  \checkmark &
  \begin{tabular}[c]{@{}l@{}}Researchers developed a framework for conceptually describing components of surveillance video, separating them into smaller components, and \\ detecting activities from some short clips of two seconds.\end{tabular} \\ \hline
\cite{Huang2020} &
  \begin{tabular}[c]{@{}l@{}}Traffic volume\\ Traffic Speed\end{tabular} &
  Deep Learning &
  Accident Detection &
  \checkmark &
  \textbf{-} &
  \textbf{-} &
  \begin{tabular}[c]{@{}l@{}}The authors present a comparative analysis of different statistical and deep learning models for solving traffic safety problems through the detection\\ of collisions and estimating crash risk in urban Interstate highways.\end{tabular} \\ \hline
\cite{Bortnikov2020} &
  \begin{tabular}[c]{@{}l@{}}Car Collision\\ Accident Detection\end{tabular} &
  Deep Learning &
  Accident Detection &
  \checkmark &
  \textbf{-} &
  \checkmark &
  \begin{tabular}[c]{@{}l@{}}Through the use of video games under different weather conditions and scene conditions, the study generated traffic data that was then processed \\ and trained with a 3D CNN. This model yielded comparable results to real-life traffic videos from YouTube.\end{tabular} \\ \hline
 \cite{Gupta2021} &
  \begin{tabular}[c]{@{}l@{}}Car Collision\\ Accident Detection\end{tabular} &
  Deep Learning &
  Accident Detection &
  \checkmark &
  \checkmark &
  \checkmark &
  Using time-dependent frames in a video, the developed model was able to evaluate the effectiveness of the model on trimmed unlabelled video. \\ \hline
\cite{Yang2021} &
  Accident Detection &
  Deep Learning &
  Accident Detection &
  \checkmark &
  \checkmark &
  \checkmark &
  \begin{tabular}[c]{@{}l@{}}In this paper, researchers propose developing a feature-fused SSD in order to improve detection accuracy of vehicles from the ImageNet video database.\end{tabular} \\ \hline
\cite{Ijjina2019} &
  \begin{tabular}[c]{@{}l@{}}Traffic Speed\\ Traffic Trajectory\\ Car Distance\end{tabular} &
  Deep Learning &
  \begin{tabular}[c]{@{}l@{}}Traffic Anomaly Detection\\ Accident Detection\end{tabular} &
  \checkmark &
  \checkmark &
  \checkmark &
  \begin{tabular}[c]{@{}l@{}}The proposed supervised deep learning framework detects and identifies road-side vehicular accidents by extracting feature points based on local \\ features such as trajectory intersection and velocity,  and by detecting anomalies in real-time accident conditions such as daylight variations.\end{tabular} \\ \hline
\cite{You2020} &
  \begin{tabular}[c]{@{}l@{}}Car Trajectory\\ Vehicle Detection\end{tabular} &
  Deep Learning &
  Accident Detection &
  \checkmark &
  \checkmark &
  \checkmark &
  \begin{tabular}[c]{@{}l@{}}In this study, the authors discovered that time segmentation methods such as SS-TCN and MS-TCN were more successful on the dataset at higher\\  IoU thresholds. In addition, the R-C3D algorithm has a comparable result when compared to segmentation-based approaches.\end{tabular} \\ \hline
\cite{Srinivasan2020} &
  Vehicle Detection &
  Deep Learning &
  \begin{tabular}[c]{@{}l@{}}Accident Detection\\ Accident Classification\end{tabular} &
  \checkmark &
  \checkmark &
  \checkmark &
  \begin{tabular}[c]{@{}l@{}}The authors developed a scalable algorithm for high-speed object detection (DETR), with a less complex architecture and a higher level of accuracy \\ compared to other object detection algorithms that are based on correlations between all objects in the video data.\end{tabular} \\ \hline
\cite{Hui2015} &
  \begin{tabular}[c]{@{}l@{}}Vehicle Trajectory\\ Traffic Speed\end{tabular} &
  Deep Learning &
  Accident Detection &
  \checkmark &
  \textbf{-} &
  \textbf{-} &
  The authors proposed using a Gaussian Mixture Model to extract foreground and background information from video streams in order to create a vision-based accident detection model. \\ \hline
\cite{Xia2018} &
  \begin{tabular}[c]{@{}l@{}}Traffic Pattern\\ Traffic Trajectory\end{tabular} &
  Probabilistic &
  Traffic Flow pattern &
  \checkmark &
  \checkmark &
  \checkmark &
  \begin{tabular}[c]{@{}l@{}}The authors applied the SIFT flow method to improve dense trajectories and generate visual words that can be utilized in detecting traffic flow.\\The data from the experiments demonstrate that the SIFT method is effective.\end{tabular} \\ \hline
\cite{Vatti2018} &
  \begin{tabular}[c]{@{}l@{}}Car Collision\\ Lane Maneuver\end{tabular} &
  Statistical Model &
  Traffic Flow Pattern &
  \textbf{-} &
  \checkmark &
  \textbf{-} &
  The authors developed an electronic notification system that can alert relatives when a vehicle accident is detected based on the vehicle's trajectory, position, and acceleration.\\ \hline
\end{tabular}%
}
\end{table*}

\subsection{RQ2:Algorithms and Taxonomies in Autonomous Transportation}

In order to answer our second research question, we have identified the most critical taxonomies and algorithms used in the AR systems for autonomous transportation and accident detection, respectively. Table~\ref{tab:my-tablerq2} shows the models, architecture, features used by other researchers, and the evaluation metrics for evaluating the performance of proposed models. It is noteworthy that most researchers employed different metrics to evaluate their algorithm's performance, especially research work that develops a new novel algorithm or benchmark. More than 60\% of the reviewed paper evaluated their algorithms using Mean Absolute Percentage Error (MAPE), Mean Absolute Error (MAE), Mean Average Precision (MAP), Intersection over Union (IOU)~\cite{nowozin2014optimal}, and Detection Rate (DR).~\cite{Yao2022} proposed a novel FOL-based method for unsupervised video anomaly detection (VAD). A metric for computing anomaly scores using the spatial-temporal area under the curve (STAUC) was introduced.~\cite{Reddy2021} developed a Spatio-Temporal Graph Neural Network for managing and predicting traffic flow, while RNN, LSTM, and other architectures were unable to fully capture it. Their study combined GNN, RNN, and a transformer layer to model complex topological and temporal relationships among traffic data, including adjacent traffic flows.~\cite{Yu} proposed a new graph-based Spatio-temporal model to predict future traffic accidents. The integration of spatial, temporal, and external features in predicting accidents achieved a performance improvement of around 5\% over the SAE.~\cite{Ali2022} developed a Graph Convolutional Network coupled with DHST-Net, called GCN-DHSTNet, which is an enhanced GCN model for learning the spatial dependence of dynamic traffic flow by applying LSTM to capture dynamic temporal correlations with other external features. In terms of RMSE and MAPE, the proposed model is 27.2\% and 11.2\% better than AAtt-DHSTNet, which is the current state of the art. \cite{Wang2020a} study focused on accident prediction that takes into account Spatio-temporal dependence and other external factors in anticipating accident occurrence. Research work done in \cite{Reddy2021} proposed a hybrid method for detecting and recognizing stationary and moving vehicles, traffic lights, and road signs using Deep Q-Learning and YOLOv3.~\cite{Bortnikov2020} study developed an HRNN for detecting accidents from CCTV surveillance by exploiting temporal and spatial features in the classification of the video footage.~\cite{Yang2021} proposed a feature-fused SSD and a new tracking-based object detection technique TDO with greatly improved detection results over state-of-the-art and also established a vehicle dataset for highway scene analysis.~\cite{Huang2019} developed a supervised learning algorithm to detect crash patterns from historical traffic data. They examined different prediction methods to estimate crash risk or occurrence.~\cite{You2020} also created a benchmark of traffic accident data based on cause and effect events with temporal intervals in each accident event. The dataset provides atomic cues for reasoning in a complex environment and planning future actions, including mitigating legal ambiguity among agents. The framework developed by~\cite{Tang2017} can classify traffic data into different categories, such as detecting a vehicle turning directions, bicycle lanes, and pedestrians within the two seconds of traffic footage. In order to correctly predict accidents and classify external factors leading to accident occurrence.~\cite{Wang2020a} take into account Spatio-temporal dependence in their proposed methodology.

\begin{table*}[!]
\caption{Identifying the main taxonomies and algorithms used in AR for autonomous transportation based on relevant studies to our second research question. The notation “-`` means the metric is not applicable.}
\label{tab:my-tablerq2}
 \renewcommand*{\arraystretch}{1.4}
\resizebox{\textwidth}{!}{%
\begin{tabular}{llllccccclccc} 
\hline
\rowcolor[rgb]{0.753,0.753,0.753} \textbf{Authors} & \textbf{Model}                                                                                                   & \textbf{Architecture}                                                                                                                    & \textbf{Model Features}                                                             & \textbf{ACC/AUC/DR/IOU}              & \textbf{RMSE/MAP/MAPE}      & \textbf{Precision}          & \textbf{F1score}            & \textbf{Recall}             & \textbf{Model Comparison}                                           & \textbf{ACC/AUC/DR/IOU}                                                                           & \textbf{MAPE/MAE}                                          & \textbf{Citation}            \\ 
\hline
\cite{Yao2022}                                        & Future Object Localization                                                                                       & \begin{tabular}[c]{@{}l@{}}Two Stream RNN \\ Ego Motion RNN\end{tabular}                                                                 & \begin{tabular}[c]{@{}l@{}}Object Localization\\ Object Detection\end{tabular}      & 73                                   & \textbf{-}& \textbf{-}& \textbf{-}& \textbf{-}& \begin{tabular}[c]{@{}l@{}}ConvAE\\ConvLSTMAE\\AnoPred\end{tabular} & \begin{tabular}[c]{@{}c@{}}64.3\\67.5\\64.8\end{tabular}                                          & \textbf{-}                               & 16                           \\ 
\hline
\cite{Yu}                                     & Deep Spatio-Temporal   Graph Convolutional Network( DSTGCN)                                                      & Graph Convolution   Network                                                                                                              & Weather, Road Network                                                               & \textbf{-}         & 0.34                        & 82                          & 85                          & 89                          & \begin{tabular}[c]{@{}l@{}}SVM\\SdAE\\ TARPML\end{tabular}          & \begin{tabular}[c]{@{}c@{}}0.79\\0.81\\0.83\end{tabular}                                          & \textbf{-}                               & 25                           \\ 
\hline
\cite{Wang2020a}      & Spatial Temporal Graph Neural Network                                                                          & Spatial-GNN + GRU + Transformer                                                                                                          & \begin{tabular}[c]{@{}l@{}}Traffic Speed\\ Spatiotemporal Dependencies\end{tabular} & \textbf{-}         & 3.99                        & \textbf{-}& \textbf{-}& \textbf{-}& \begin{tabular}[c]{@{}l@{}}FC-LSTM\\STGNN\end{tabular}              & \textbf{-}                                                                      & \begin{tabular}[c]{@{}c@{}}3.44\\2.62\end{tabular}         & 82                           \\ 
\hline
\cite{Bao2020}   & Spatio Temporal GCN (Graph Convolution Network)                                                                & Graph Convolution +RNN                                                                                                                   & Accident Relevant Cues                                                              & \textbf{-}         & 53.7                        & \textbf{-}& \textbf{-}& \textbf{-}& \begin{tabular}[c]{@{}l@{}}adaLEA\\DSA\end{tabular}                 & \begin{tabular}[c]{@{}c@{}}\textbf{-}\\ \textbf{-}\end{tabular} & \begin{tabular}[c]{@{}c@{}}52.3\\48.1\end{tabular}         & 15                           \\ 
\hline
 \cite{Reddy2021}& Deep Q-Learning & \begin{tabular}[c]{@{}l@{}}Deep Q-Learning\\ YOLOv3\end{tabular}                                  & \begin{tabular}[c]{@{}l@{}}Car Speed\\ Distance and Position\end{tabular}           & 90                                   & \textbf{-}& 87                          & \textbf{-}& \textbf{-}& Yolo 3                                                              & 85                                                                                                & \textbf{-}                               & \textbf{-} \\ 
\hline
\cite{Fernandez-Llorca2020}                              & Two-Stream Network                                                                                               & \begin{tabular}[c]{@{}l@{}}Two- Stream Convolutio\textbackslash{}textbf\{-\}l Networks\\ Spatiotemporal Multiplier Networks\end{tabular} & Lane Change                                                                         & 90.3                                 & \textbf{-}& \textbf{-}& \textbf{-}& \textbf{-}& Disjoint Two-Stream Convolution                                     & 89.6                                                                                              & \textbf{-}                               & 7                            \\ 
\hline
\cite{Ali2022}                       & Dynamic deep   spatio-temporal neural network (DHSTNet)                                                          & Graph Convolution Network + LSTM                                                                                                         & \begin{tabular}[c]{@{}l@{}}Weather Condition\\ Traffic Flow\end{tabular}            & \textbf{-}         & 11.08                       & \textbf{-}& \textbf{-}& \textbf{-}& \begin{tabular}[c]{@{}l@{}}DHSTNet\\ Aatt-DHSTNet\end{tabular}      & \textbf{-}                                                                      & \begin{tabular}[c]{@{}c@{}}12.80\\ 13.72\end{tabular}      & 4                            \\ 
\hline
\cite{Wang2020}                                     & \begin{tabular}[c]{@{}l@{}}Spatial-Temporal Mixed Attention \\Graph-based Convolution model (STMAG)\end{tabular} & GRU+Mixed Attention Mechanism                                                                                                        & \begin{tabular}[c]{@{}l@{}}Object Detection\\ Lane Marking\end{tabular}             & \textbf{-}         & 3.23                        & \textbf{-}& \textbf{-}& \textbf{-}& \begin{tabular}[c]{@{}l@{}}XGBOOST\\ SVR\\ LSTM\end{tabular}        & \textbf{-}                                                                      & \begin{tabular}[c]{@{}c@{}}3.71\\ 3.99\\ 3.43\end{tabular} & 7                            \\ 
\hline
\cite{Huang2020}                                       & CNN-Traffic Incident   Management (TIM)                                                                          & CNN                                                                                                                                      & \begin{tabular}[c]{@{}l@{}}Car Speed\\ Road Occupancy\end{tabular}                  & 80                                   & \textbf{-}& \textbf{-}& 78                          & \textbf{-}& RF                                                                  & 76                                                                                                & \textbf{-}                               & 31                           \\ 
\hline
 \cite{Bortnikov2020}                                & 3D Convolutional   Neural Network (CNN)                                                                          & CNN                                                                                                                                      & \begin{tabular}[c]{@{}l@{}}Optical Flow\\ Vehicle Trajectory\end{tabular}           & \textbf{-} & \textbf{-}& \textbf{-}& 71                          & \textbf{-}& \textbf{-}                                        & \textbf{-}                                                                      & \textbf{-}                               & 13                           \\ 
\hline
 \cite{Gupta2021}                                    & Time-Distributed   RNN                                                                                           & LSTM                                                                                                                                     & \begin{tabular}[c]{@{}l@{}}Temporal Features\\ Hierarchical Features\end{tabular}   & 94                                   & \textbf{-}& 95                          & 84                          & 75                          & \textbf{-}                                        & \textbf{-}                                                                      & \textbf{-}                               & 1                            \\ 
\hline
\cite{Yang2021}& Feature-Fused SSD Detector                                                                                       & Single Shot MultiBox Detector (SSD)                                                                                                      & Detection box                                                                       & \textbf{-}         & 70.5                        & \textbf{-}& \textbf{-}& \textbf{-}& \begin{tabular}[c]{@{}l@{}}SSD\\ TPN\end{tabular}                   & \textbf{-}                                                                      & \textbf{-}                               & 2                            \\ 
\hline
\cite{Ijjina2019}  & Mask R-CNN                                                                                                       & Deep CNN                                                                                                                                 & \begin{tabular}[c]{@{}l@{}}Car Speed\\ Vehicle Trajectory\end{tabular}              & DR- 71                               & \textbf{-}& \textbf{-}& \textbf{-}& \textbf{-}& Deep Spatio Temporal Network                                        & DR-77                                                                                             & \textbf{-}                               & 24                           \\ 
\hline
\cite{You2020}                                         & Single-Stream Temporal   Action Proposals (SST)                                                                  & Temporal Segment Networks (TSN)                                                                                                          & \textbf{-}                                                        & IOU - 42.07                          & \textbf{-}& \textbf{-}& \textbf{-}& \textbf{-}& \begin{tabular}[c]{@{}l@{}}R-C3D\\ MS-TCN\end{tabular}              & \textbf{-}                                                                      & \textbf{-}                               & 10                           \\ 
\hline
 \cite{Srinivasan2020}                                 & Detection Transformers   and Random Forest Classifier (DETR)                                                     & DETR + CNN                                                                                                                               & Object Detection                                                                    & 78.7                                 & \textbf{-}& 77                          & 77                          & 78                          & \begin{tabular}[c]{@{}l@{}}ARRS\\ CVABTS\end{tabular}               & \begin{tabular}[c]{@{}c@{}}DR- 50\\ DR- 71\end{tabular}                                           & \textbf{-}                               & 1                            \\ 
\hline
 \cite{Hui2015}                                  & Gaussian Mixture Model   (GMM)                                                                                   & \textbf{-}                                                                                                             & \begin{tabular}[c]{@{}l@{}}Vehicle Detection\\ Object Tracking\end{tabular}         & \textbf{-}         & \textbf{-}& \textbf{-}& \textbf{-}& \textbf{-}& \textbf{-}                                        & \textbf{-}                                                                      & \textbf{-}                               & 39                           \\ 
\hline
\cite{Xia2018}                                         & Sparse Topic Model                                                                                               & \begin{tabular}[c]{@{}l@{}}Scale-invariant Feature Transform\\ (SIFT) flow\end{tabular}                                                  & Motion Pattern                                                                      & AUC - 91.2                           & \textbf{-}& \textbf{-}& \textbf{-}& \textbf{-}& \begin{tabular}[c]{@{}l@{}}GPR\\JSM\\ BiLSTM\end{tabular}           & \begin{tabular}[c]{@{}c@{}}85.5\\ 80.2\\ 88.1\end{tabular}                                        & \textbf{-}                               & 5                            \\ 
\hline
\cite{Vatti2018}                                    & Accident Detection and   Communication System                                                                    & \textbf{-}                                                                                                             & Motion Pattern                                                                      & \textbf{-}         & \textbf{-}& \textbf{-}& \textbf{-}& \textbf{-}& \textbf{-}                                        & \textbf{-}                                                                      & \textbf{-}                               & 16                           \\
\hline
\end{tabular}}
\end{table*}


\subsection{RQ3: Main Dataset, Features and Metrics Used in Action Recognition for Accident Detection Task}

Our third research question focused on exploring the dataset used for accident detection. Table \ref{tab:my-tablerq3} showcase the dataset features, type of sensors/video data, and the link to publicly available datasets for accident detection in a smart city.
~\cite{Yao2022} developed a benchmark dataset to assess the quality of traffic accident detection and anomaly detection for nine action classes. Based on the scarcity of annotated real-life accident datasets,~\cite{Bortnikov2020} utilized simulated game video data with varied weather and scene conditions and yielded comparable results to real-life traffic videos on YouTube, as shown in Table~\ref{tab:my-tablerq3}. The majority of the dataset used in accident detection and autonomous vehicles are collected from dashcams, traffic surveillance cameras, drones such as HighD, InD, or Interaction datasets~\cite{krajewski2018highd, zhan2019interaction} and cameras installed on buildings. For example, NGSIM HW101 and NGSIM I-80 datasets~\cite{colyar2007ngsim,halkias2006ngsim} contain 45 minutes of images recorded from a building for eight synchronized cameras at 10 Hz.~\cite{Fernandez-Llorca2020} suggest that this dataset (NGSIM HW101) is not fully applicable for onboard detection applications even though it is beneficial for understanding and assessing the motion and behavior of vehicles and drivers under different traffic conditions. \href{http://www.poss.pku.edu.cn/download}{PKU dataset} includes more than 5700 environmental trajectories collected using multiple horizontal 2-D lidars covering $\mathrm{360^\circ}$, including vehicles' trajectory data over 64 km and 19 hours of footage~\cite{zhao-pku}. The \href{https://prevention-dataset.uah.es/}{Prevention dataset} includes data from three radars, two cameras, and one light detection and ranging (LiDAR), covering a range of 80 meters around an ego-vehicle, to support the development of intelligent systems for vehicle detection and tracking \cite{izquierdo2019prevention}. In a similar fashion, the \href{http://apolloscape.auto/#}{apolloscape dataset} was developed to support automatic driving and navigation in smart cities. The dataset contains about 100K image frames and 1000km trajectories collected using four cameras and two laser scanners utilizing 3D perception LiDAR object detection, and tracking \cite{wang2019apolloscape}. \cite{Ijjina2019} compiled surveillance videos at 30 frames per second (FPS) trimmed down to 20 seconds video chunks collected from CCTV videos recorded at road intersections from different parts of the world with diversified ambient conditions such as harsh sunlight, daylight hours, snow and night hours.

\begin{table*}
\caption{Overview of datasets used in AR for autonomous transportation, features of the datasets and download links to the datasets.}
\label{tab:my-tablerq3}
 \renewcommand*{\arraystretch}{1.4}
\resizebox{\textwidth}{!}
{%
\begin{tabular}{lllllll}
\hline
\rowcolor[HTML]{C0C0C0} 
\textbf{Authors} &
  \textbf{Model} &
  \textbf{Dataset} &
  \textbf{Fps} &
  \textbf{Dataset Feature} &
  \textbf{Accesibility} &
  \textbf{Data Collection Approach} \\ \hline
\cite{Yao2022} &
  Future Object   localization &
  DoTA (Detection of Traffic Anomaly) &
  10 &
  4,677 videos &
  Yes-\href{https://github.com/MoonBlvd/Detection-of-Traffic-Anomal}{link} &
  Dashcam \\ \hline
\cite{Yu} &
  Deep Spatio-Temporal Graph Convolutional Network ( DSTGCN) &
  \begin{tabular}[l]{@{}l@{}}Traffic Accident Data, Taxi GPS Data\\Meteorological data\end{tabular} &\textbf{-}
   &
  \begin{tabular}[l]{@{}l@{}}Varied weather (Cloudy, Snow, etc.)\\ Road network and PoI (Point of interest)\end{tabular} &
 NA&
  \begin{tabular}[l]{@{}l@{}}Sensors\\ Traffic Surveillance\end{tabular} \\ \hline
\cite{Wang2020a} &
  Spatial Temporal Graph Neural Network &
  \begin{tabular}[c]{@{}c@{}}PEMS-BAY\\ META-LA\end{tabular} &\textbf{-}
   &
  \begin{tabular}[l]{@{}l@{}}53116 videos from 325 sensors\\ 34272   videos from 207 sensors\end{tabular} &
  Yes-\href{https://pems.dot.ca.gov/}{link} &
  \begin{tabular}[l]{@{}l@{}}Sensors\\ Traffic Surveillance\end{tabular} \\ \hline

\cite{Bao2020} &
  Spatio Temporal GCN (Graph Convolution Network) &
  \begin{tabular}[l]{@{}l@{}}CCD (Car Crash Dataset)\\DAD and A3D\end{tabular} &\textbf{-}
  &
  \begin{tabular}[l]{@{}l@{}}6.35 hours\\ Varied   weather conditions (snow, day and night)\\  2.43 hours and 3.56 hours\end{tabular} & 
  Yes-\href{https://github.com/Cogito2012/UString}{link}
&  Dashcam \\ \hline

\cite{Reddy2021} &
  Deep Q-Learning &
  Traffic   Driving data &\textbf{-}
  &
  182 drive sequences &
 NA&
  Dashcam \\ \hline

 \cite{Fernandez-Llorca2020} &
  Two-Stream Network &
  PREVENTION dataset &\textbf{-}
  &
  \begin{tabular}[l]{@{}l@{}}6 hours video (80 meters around ego-vehicle)\\ 3 radars 2 cameras and 1 LiDAR\end{tabular} &
  Yes-\href{https://prevention-dataset.uah.es/}{link} &
  \begin{tabular}[l]{@{}l@{}}LiDAR Rada\\ DashCam\end{tabular} \\ \hline
\cite{Ali2022} &
  Dynamic deep spatio-temporal neural network (DHSTNet) &
  \begin{tabular}[l]{@{}l@{}}TaxiBj\\ Bike NYC\end{tabular} &\textbf{-}&
  16 months video recordings &
 NA&
  Sensors \\ \hline
 \cite{Wang2020} &
  Spatial-Temporal Mixed   Attention Graph-based Convolution model (STMAG) &
  Curated Traffic Data &\textbf{-}
  &
  2000 videos &
  No-Future Release &
  Dashcam \\ \hline
 \cite{Alkandari2015} &
  Dynamic Webster with   dynamic Cycle Time algorithm (DWDC) &\textbf{-}&
\textbf{-}&
 \textbf{-}&
 NA&
  Sensors \\ \hline
 \cite{Riaz2022} &
  FWPredNet &
  \begin{tabular}[l]{@{}l@{}}KITTI\\ HTA\\ D2City\end{tabular} &
  65 &
  \begin{tabular}[l]{@{}l@{}}600 frames from Kitti\\ 286 clips\\ 65 frames and 678 video\end{tabular} &
  Yes-\href{https://github.com/MoonBlvd/Detection-of-Traffic-Anomal}{link} &
  Dashcam \\ \hline
 \cite{Wang2020a} &
  Mixure LDA and   Expectation–maximization algorithm &
  40- seconds traffic video &
  15 &
  600*800 frame dimension &
 NA&
  Traffic Surveillance \\ \hline
  \cite{Huang2020} &
  CNN-Traffic Incident   Management (TIM) &
  \begin{tabular}[l]{@{}l@{}}Traffic Management Centers report\\  IOWA DOT radar sensors\end{tabular} &\textbf{-}
  &
  \begin{tabular}[l]{@{}l@{}}856 crash reports\\ 29   sensors\end{tabular} &
  \begin{tabular}[l]{@{}l@{}}Yes- \href{https://mesonet.agron.iastate.edu/request/rwis/traffic.phtml\\ }{link}\end{tabular} &
  Sensors \\ \hline
\cite{Bortnikov2020} &
  3D Convolutional   Neural Network (CNN) &
  \begin{tabular}[l]{@{}l@{}}Video game GTA V\\ YouTube Car Accident Video\end{tabular} &\textbf{-}
  &
  5 hours recording &
 NA&
  \begin{tabular}[l]{@{}l@{}}Traffic Surveillance\\ Dashcam\end{tabular} \\ \hline
  \cite{Gupta2021} &
  Time-distributed   RNN &
  DETRAC dataset &
  25 &
  \begin{tabular}[l]{@{}l@{}}10 hours recording of 376   videos\\  99 frames selected from each video\end{tabular} &
  \begin{tabular}[l]{@{}l@{}}Yes-\href{https://detrac-db.rit.albany.edu/download\\ }{link}\end{tabular} &
  Traffic Surveillance \\ \hline

  \cite{Yang2021} &
  Feature-fused SSD   Detector&
  \begin{tabular}[l]{@{}l@{}}Highway vehicle dataset\\ ImageNet VID dataset\end{tabular} &
  \textbf{-}&
  \begin{tabular}[l]{@{}l@{}}32938 vehicle samples\\ 5354 videos\end{tabular} &
  Yes-\href{https://image-net.org/download.php}{link} &
  Traffic Surveillance \\ \hline
 \cite{Ijjina2019} &
  Mask R-CNN &
  YouTube Accident Videos &
  30 &
  \begin{tabular}[l]{@{}l@{}}20 seconds video chunks\\ Varied weather (harsh sunlight, daylight hours)\end{tabular} &
 NA&
  Traffic Surveillance \\ \hline
 \cite{You2020} &
  Single-Stream Temporal   Action Proposals (SST) &
  Causality in Traffic Accident (CTA) &\textbf{-}
  &
  \begin{tabular}[l]{@{}l@{}}9.53 hours video from 1935 videos.\\  18 semantic cause and 7 semantic effect labels\end{tabular} &
  Yes-\href{https://github.com/tackgeun/CausalityInTrafficAccident}{link} &
  Dashcam \\ \hline
\cite{Srinivasan2020} &
  Detection Transformers   and Random Forest Classifier (DETR) &
  CADP &\textbf{-}
  &
  1416 Accident footage &
  Yes-\href{https://ankitshah009.github.io/accident\_forecasting\_traffic\_camera}{link} &
  Traffic Surveillance \\ \hline
 \cite{Hui2015} &
  Gaussian Mixture Model   (GMM) &
 \textbf{-}&
 \textbf{-}&
 \textbf{-}&
 NA&Sensors
  \\ \hline

 \cite{Xia2018} &
  Sparse Topic Model &
  \begin{tabular}[l]{@{}l@{}}QMUL Junction dataset\\ AVSS dataset\end{tabular} &
  25 &
  \begin{tabular}[l]{@{}l@{}}52 mins traffic video\\ 4 seconds chunks\end{tabular} &
  \begin{tabular}[l]{@{}l@{}}Yes-\href{https://personal.ie.cuhk.edu.hk/$\sim$ccloy/downloads\_qmul\_junction.html\\ }{link}\end{tabular} &
  Traffic Surveillance \\ \hline
  \cite{Vatti2018} &
  Accident Detection and   Communication System &
\textbf{-}&
\textbf{-}&
\textbf{-}&
 NA& Sensors \\ \hline
\end{tabular}}
\end{table*}



\section{Limitation}\label{limitation}
Our research focused on research papers relevant to action recognition, accident detection, and autonomous transportation published within the last ten years that developed a novel framework or benchmark dataset. According to our inclusion and exclusion criteria, we excluded research papers in languages other than English and those that did not include video/motion analysis. Consequently, only a fraction of the articles surveyed in the study were considered. The vast majority of AR techniques developed in another domain can also easily be applied in a new domain (such as Accident Detection). In light of this, it is recommended to conduct a research project that combines Action Recognition Techniques for objects and human action classification since both have been developed using similar model architectures.

\section{Conclusion}\label{conclusion}
This systematic literature review aims to determine state-of-the-art Action Recognition for accident detection and autonomous transportation in smart cities. In order to achieve this, we used the PRISMA guideline for selecting seminary articles related to our topic domain. This guideline was based on the Inclusion and Exclusion criteria discussed in Section \ref{Methods}. We selected 22 papers from an initial list of 2030 publications, and we categorized and analyzed the relevant literature based on the three pillars of our research question. This paper discussed the leading techniques and applications of action recognition in autonomous transportation. The study also explored the main taxonomies and algorithms used in AR for autonomous transportation. Finally, we presented an overview of datasets used in AR for autonomous transportation, features of the datasets, and download links to the datasets.

In the quest for a smart city, automating city traffic by capturing spatial and temporal information from DNN is a significant step in smart city automation.~\cite{Bao2020} developed a model to handle the challenges of relational feature learning and uncertainty anticipation from traffic video to predict accident occurrence within 3.53 seconds with an average precision of 72.22\% using Graph Convolution Network (GCN) and Bayesian Neural Networks (BNN). Many factors are involved in traffic accidents, including driver behavior, weather conditions, traffic flow, and road structure.~\cite{Yu} examined spatial-temporal relationships on heterogeneous data to develop a road-level accident prediction system. Besides sequential patterns in the temporal dimension, traffic flows on the road are strongly affected by other road networks in the spatial dimension. Studies have been conducted on traffic flow prediction; however, many of them lack the ability to account for spatial and temporal dependencie~\cite{Wang2020a}.~\cite{Reddy2021} aimed to extract roadway characteristics that are relevant to the trajectory of an autonomous vehicle from real-world road conditions using Deep Q-Learning. Analyzing and forecasting dynamic traffic patterns within smart cities are necessary for planning and managing transportation. Forecasting traffic flow has become more difficult because of the volatility of vehicle flow in the temporal dimension and the uncertainty related to accident occurrence and traffic movements.~\cite{Ali2022} proposed a hybrid model composed of GCN and DHSTNet, which can forecast short-term traffic patterns in urban areas for improved traffic management. Similarly,~\cite{Alkandari2015} developed a methodology for determining how long a vehicle stays in traffic based on traffic flow and congestion.

Automation of accident detection using AI systems based on security cameras will be a step towards the security of more lives. It will also support the transformation of traffic cameras to support smart city automation and provide first responders and law enforcement agencies with information about road accidents. We recommend future research focus on scaling up accident detection systems that can be integrated into smart city automation for alerting first responders about road accidents and providing a quick response to victims thereby reducing human error and response time by adopting a spontaneous model for reporting accidents.
\section*{Acknowledgment}
The authors would thank Annu Prabhakar for her recommendations regarding writing a systematic review. Also, we would like to thank Sylvia Azumah, Jones Yeboah, and Izunna Okpala for their help reviewing the search results and selecting papers based on the inclusion and exclusion criteria.

\section*{Funding statement}
This research did not receive any specific grant from funding
agencies in the public, commercial, or not-for-profit sectors.

\section*{Declaration of Competing Interest}
The authors declare that there are no conflicts of interest.


\bibliographystyle{cas-model2-names}

\bibliography{cas-refs}





\end{document}